\documentclass[11pt,a4paper]{article}
\usepackage[hyperref]{acl2018}
\usepackage{times}
\usepackage{latexsym}
\usepackage{graphicx}
\usepackage{amsmath,amssymb}
\usepackage{caption}
\usepackage{adjustbox}
\usepackage{makecell}
\usepackage{float}

\usepackage{url}

\aclfinalcopy 
\def\aclpaperid{1397} 

\title{Conditional Generators of Words Definitions}

\author{Artyom Gadetsky \\
  National Research University \\
  Higher School of Economics \\
  {\tt artygadetsky@yandex.ru} \\\And
  Ilya Yakubovskiy \\
  Joom \\
  {\tt yakubovskiy@joom.com} \\\And
  Dmitry Vetrov \\
  National Research University \\
  Higher School of Economics \\
  Samsung-HSE Laboratory \\
  {\tt vetrovd@yandex.ru}
}

\date{}

\begin{document}
\maketitle
\begin{abstract}
We explore recently introduced definition modeling technique that provided the tool for evaluation of different distributed vector representations of words through modeling dictionary definitions of words. In this work,  we study the problem of word ambiguities in definition modeling and propose a possible solution by employing latent variable modeling and soft attention mechanisms. Our quantitative and qualitative evaluation and analysis of the model shows that taking into account words’ ambiguity and polysemy leads to performance improvement.
\end{abstract}

\section{Introduction}
Continuous representations of words are used in many natural language processing (NLP) applications. Using pre-trained high-quality word embeddings are most effective if not millions of training examples are available, which is true for most tasks in NLP \citep{pmlr-v48-kumar16, karpathy2015deep}. Recently, several unsupervised methods were introduced to learn word vectors from large corpora of texts \cite{word2vec, glove, joulin2016fasttext}.
Learned vector representations have been shown to have useful and interesting properties. For example, Mikolov et al. \shortcite{word2vec} showed that vector operations such as subtraction or addition reflect semantic relations between words. Despite all these properties it is hard to precisely evaluate embeddings because analogy relation or word similarity tasks measure learned information indirectly.

Quite recently Noraset et al. \shortcite{noraset2016definition} introduced a more direct way for word embeddings evaluation. Authors suggested considering definition modeling as the evaluation task. In definition modeling vector representations of words are used for conditional generation of corresponding word definitions. The primary motivation is that high-quality word embedding should contain all useful information to reconstruct the definition. The important drawback of \citet{noraset2016definition} definition models is that they cannot take into account words with several different meanings. These problems are related to word disambiguation task, which is a common problem in natural language processing. Such examples of polysemantic words as ``bank`` or ``spring`` whose meanings can only be disambiguated using their contexts. In such cases, proposed models tend to generate definitions based on the most frequent meaning of the corresponding word. Therefore, building models that incorporate word sense disambiguation is an important research direction in natural language processing.

In this work, we study the problem of word ambiguity in definition modeling task. We propose several models which can be possible solutions to it. One of them is based on recently proposed Adaptive Skip Gram model \citep{bartunov2016breaking}, the generalized version of the original SkipGram Word2Vec, which can differ word meanings using word context. The second one is the attention-based model that uses the context of a word being defined to determine components of embedding referring to relevant word meaning. Our contributions are as follows: (1) we introduce two models based on recurrent neural network (RNN) language models, (2) we collect new dataset of definitions, which is larger in number of unique words than proposed in \citet{noraset2016definition} and also supplement it with examples of the word usage (3) finally, in the experiment section we show that our models outperform previously proposed models and have the ability to generate definitions depending on the meaning of words.
\section{Related Work}

\subsection{Constructing Embeddings Using Dictionary Definitions}
Several works utilize word definitions to learn embeddings. For example, \citet{TACL711} use definitions to construct sentence embeddings. Authors propose to train recurrent neural network producing an embedding of the dictionary definition that is close to an embedding of the corresponding word. The model is evaluated with the reverse dictionary task. \citet{bahdanau2017learning} suggest using definitions to compute embeddings for out-of-vocabulary words. In comparison to \citet{TACL711} work, dictionary reader network is trained end-to-end for a specific task.

\subsection{Definition Modeling}
Definition modeling was introduced in \citet{noraset2016definition} work. The goal of the definition model $p(D|w^*)$ is to predict the probability of words in the definition $D = \{w_1, \dots, w_T\}$ given the word being defined $w^*$. The joint probability is decomposed into separate conditional probabilities, each of which is modeled using the recurrent neural network with soft-max activation, applied to its logits.
\begin{equation} \label{eq:1}
    p(D | w^*) = \prod_{t=1}^{T}p(w_t | w_{i < t}, w^*)
\end{equation}

Authors of definition modeling consider following conditional models and their combinations: \textit{Seed (S)} - providing word being defined at the first step of the RNN, \textit{Input (I)} - concatenation of embedding for word being defined with embedding of word on corresponding time step of the RNN, \textit{Gated (G)}, which is the modification of GRU cell. Authors use a character-level convolutional neural network (CNN) to provide character-level information about the word being defined, this feature vector is denoted as \textit{(CH)}. One more type of conditioning referred to as \textit{(HE)}, is hypernym relations between words, extracted using Hearst-like patterns.

\section{Word Embeddings}

Many natural language processing applications treat words as atomic units and represent them as continuous vectors for further use in machine learning models. Therefore, learning high-quality vector representations is the important task.

\subsection{Skip-gram}

One of the most popular and frequently used vector representations is Skip-gram model. The original Skip-gram model consists of grouped word prediction tasks. Each task is formulated as a prediction of the word $v$ given word $w$ using their input and output representations:
\begin{equation} \label{eq:2}
p(v|w, \theta) = \frac{exp(in^T_w out_v)}{\sum_{v'=1}^{V} \exp(in^T_w out_{v'})}
\end{equation}
where $\theta$ and $V$ stand for the set of input and output word representations, and dictionary size respectively.
These individual prediction tasks are grouped in a way to independently predict all adjacent (with some sliding window) words $y = \{y_1, \dots y_C \}$ given the central word $x$: 
\begin{equation} \label{eq:3}
p(y | x, \theta) = \prod_{j} p(y_j | x, \theta)
\end{equation}
The joint probability of the model is written as follows:
\begin{equation} \label{py_of_x}
p(Y|X, \theta) = \prod_{i=1}^{N} p(y_i | x_i, \theta)
\end{equation}
where $(X, Y) = \{x_i, y_i\}_{i=1}^{N}$ are training pairs of words and corresponding contexts and $\theta$ stands for trainable parameters.

Also, optimization of the original Skip-gram objective can be changed to a negative sampling procedure as described in the original paper or hierarchical soft-max prediction model \cite{hsoftmax} can be used instead of (\ref{eq:2}) to deal with computational costs of the denominator. After training, the input representations are treated as word vectors. 

\subsection{Adaptive Skip-gram}

Skip-gram model maintains only one vector representation per word that leads to mixing of meanings for polysemantic words. \citet{bartunov2016breaking} propose a solution to the described problem using latent variable modeling. They extend Skip-gram to Adaptive Skip-gram (AdaGram) in a way to automatically learn the required number of vector representations for each word using Bayesian nonparametric approach. In comparison with Skip-gram AdaGram assumes several meanings for each word and therefore keeps several vectors representations for each word. They introduce latent variable $z$ that encodes the index of meaning and extend (\ref{eq:2}) to $p(v|z, w, \theta)$. They use hierarchical soft-max approach rather than negative sampling to overcome computing denominator. 
\begin{equation} \label{ada1}
p(v|z=k, w, \theta) = \prod_{n \in path(v)} \sigma(ch(n) in_{wk}^T out_n)
\end{equation}
Here $in_{wk}$ stands for input representation of word $w$ with meaning index $k$ and output representations are associated with nodes in a binary tree, where leaves are all possible words in model vocabulary with unique paths from the root to the corresponding leaf. $ch(n)$ is a function which returns 1 or -1 to each node in the $path(\cdot)$ depending on whether $n$ is a left or a right child of the previous node in the path. Huffman tree is often used for computational efficiency.

To automatically determine the number of meanings for each word authors use the constructive definition of Dirichlet process via stick-breaking representation ($p(z=k|w, \beta)$), which is commonly used prior distribution on discrete latent variables when the number of possible values is unknown (e.g. infinite mixtures). 
\begin{equation} \label{ada2}
\begin{split}
& p(z=k | w, \beta) = \beta_{wk} \prod_{r=1}^{k-1}(1-\beta_{wr}) \\
& p(\beta_{wk} | \alpha) = Beta(\beta_{wk} | 1, \alpha)
\end{split}
\end{equation}

This model assumes that an infinite number of meanings for each word may exist. Providing that we have a finite amount of data, it can be shown that only several meanings for each word will have non-zero prior probabilities.

Finally, the joint probability of all variables in AdaGram model has the following form:
\begin{equation} \label{ada3}
\begin{split}
   & p(Y, Z, \beta | X, \alpha, \theta) = \prod_{w=1}^{V} \prod_{k=1}^{\infty} p(\beta_{wk} | \alpha) \cdot\\
    & \cdot \prod_{i=1}^{N} [p(z_i | x_i, \beta) \prod_{j=1}^{C} p(y_{ij} | z_i, x_i, \theta)]
\end{split}
\end{equation}
Model is trained by optimizing Evidence Lower Bound using stochastic variational inference \citep{svi} with fully factorized variational approximation of the posterior distribution $p(Z, \beta | X, Y, \alpha, \theta) \approx q(Z) q(\beta)$.

One important property of the model is an ability to disambiguate words using context. More formally, after training on data $D=\{x_i, y_i\}_{i=1}^{N}$ we may compute the posterior probability of word meaning given context and take the word vector with the highest probability.:
\begin{equation} \label{disamb}
\begin{split}
& p(z=k | x, y, \theta) \propto \\
& \propto p(y|x, z=k, \theta) \int p(z=k|\beta, x) q(\beta) d\beta \\
\end{split}
\end{equation}
This knowledge about word meaning will be further utilized in one of our models as $disambiguation(x|y)$.

\section{Models}

In this section, we describe our extension to original definition model. The goal of the extended definition model is to predict the probability of a definition $D = \{w_1, \dots, w_T\}$ given a word being defined $w^*$ and its context $C=\{c_1, \dots, c_m \}$ (e.g. example of use of this word). As it was motivated earlier, the context will provide proper information about word meaning. The joint probability is also decomposed in the conditional probabilities, each of which is provided with the information about context:
\begin{equation} \label{eq:4}
    p(D | w^*, C) = \prod_{t=1}^{T}p(w_t | w_{i < t}, w^*, C)
\end{equation}

\subsection{AdaGram based}

Our first model is based on original \textit{Input (I)} conditioned on Adaptive Skip-gram vector representations. To determine which word embedding to provide as \textit{Input (I)} we disambiguate word being defined using its context words $C$. More formally our \textit{Input (I)} conditioning is turning in:
\begin{equation} \label{eq:5}
\begin{split}
    h_t &= g([v^*; v_{t}], h_{t-1}) \\
    v^* &= disambiguation(w^* | C)
\end{split}
\end{equation}

where $g$ is the recurrent cell, $[a; b]$ denotes vector concatenation, $v^{*}$ and $v_t$ are embedding of word being defined $w$ and embedding of word $w_t$ respectively. We refer to this model as \textit{Input Adaptive (I-Adaptive)}.

\subsection{Attention based}

Adaptive Skip-gram model is very sensitive to the choice of concentration parameter in Dirichlet process. The improper setting will cause many similar vectors representations with smoothed meanings due to theoretical guarantees on a number of learned components. To overcome this problem and to get rid of careful tuning of this hyper-parameter we introduce following model:
\begin{equation} \label{eq:6}
\begin{split}
    & h_t = g([a^*; v_{t}], h_{t-1}) \\
    & a^* = v^* \odot mask \\
    & mask = \sigma(W \frac{\sum_{i=1}^{m} ANN(c_i)}{m} + b)
\end{split}
\end{equation}

where $\odot$ is an element-wise product, $\sigma$ is a logistic sigmoid function and $ANN$ is attention neural network, which is a feed-forward neural network. We motivate these updates by the fact, that after learning Skip-gram model on a large corpus, vector representation for each word will absorb information about every meaning of the word. Using soft binary mask dependent on word context we extract components of word embedding relevant to corresponding meaning. We refer to this model as \textit{Input Attention (I-Attention)}.

\subsection{Attention SkipGram}

For attention-based model, we use different embeddings for context words. Because of that, we pre-train attention block containing embeddings, attention neural network and linear layer weights by optimizing a negative sampling loss function in the same manner as the original Skip-gram model:

\begin{equation} \label{eq:7}
\begin{split}
    & \log \sigma(v'^{T}_{w_O} v_{w_I}) \\
    & + \sum_{i=1}^{k} \mathbb{E}_{w_i \sim P_n(w)} [ \log \sigma(-v'^T_{w_i}v_{w_I})]
\end{split}
\end{equation}

where $v'_{w_O}$, $v_{w_I}$ and $v'_{w_i}$ are vector representation of "positive" example, anchor word and negative example respectively. Vector $v_{w_I}$ is computed using embedding of $w_I$ and attention mechanism proposed in previous section.

\section{Experiments}

\subsection{Data}

\begin{table}
\centering
\begin{tabular}{  r ||  r | r | r }
 \hline
 Split & train & val & test\\
 \hline
 \#Words   & 33,128 & 8,867 & 8,850\\
 \#Entries &  97,855  & 12,232 & 12,232\\
 \#Tokens & 1,078,828 & 134,486 & 133,987\\
 Avg length    & 11.03 & 10.99 &  10.95\\
 \hline
\end{tabular}
\captionof{table}{Statistics of new dataset}
\label{dataset}
\end{table}

We collected new dataset of definitions using \citet{oxford} API. Each entry is a triplet, containing the word, its definition and example of the use of this word in the given meaning. It is important to note that in our data set words can have one or more meanings, depending on the corresponding entry in the Oxford Dictionary. Table \ref{dataset} shows basic statistics of the new dataset.

\subsection{Pre-training}
\begin{figure}
  \centering
  \includegraphics[keepaspectratio, width=0.45\textwidth]{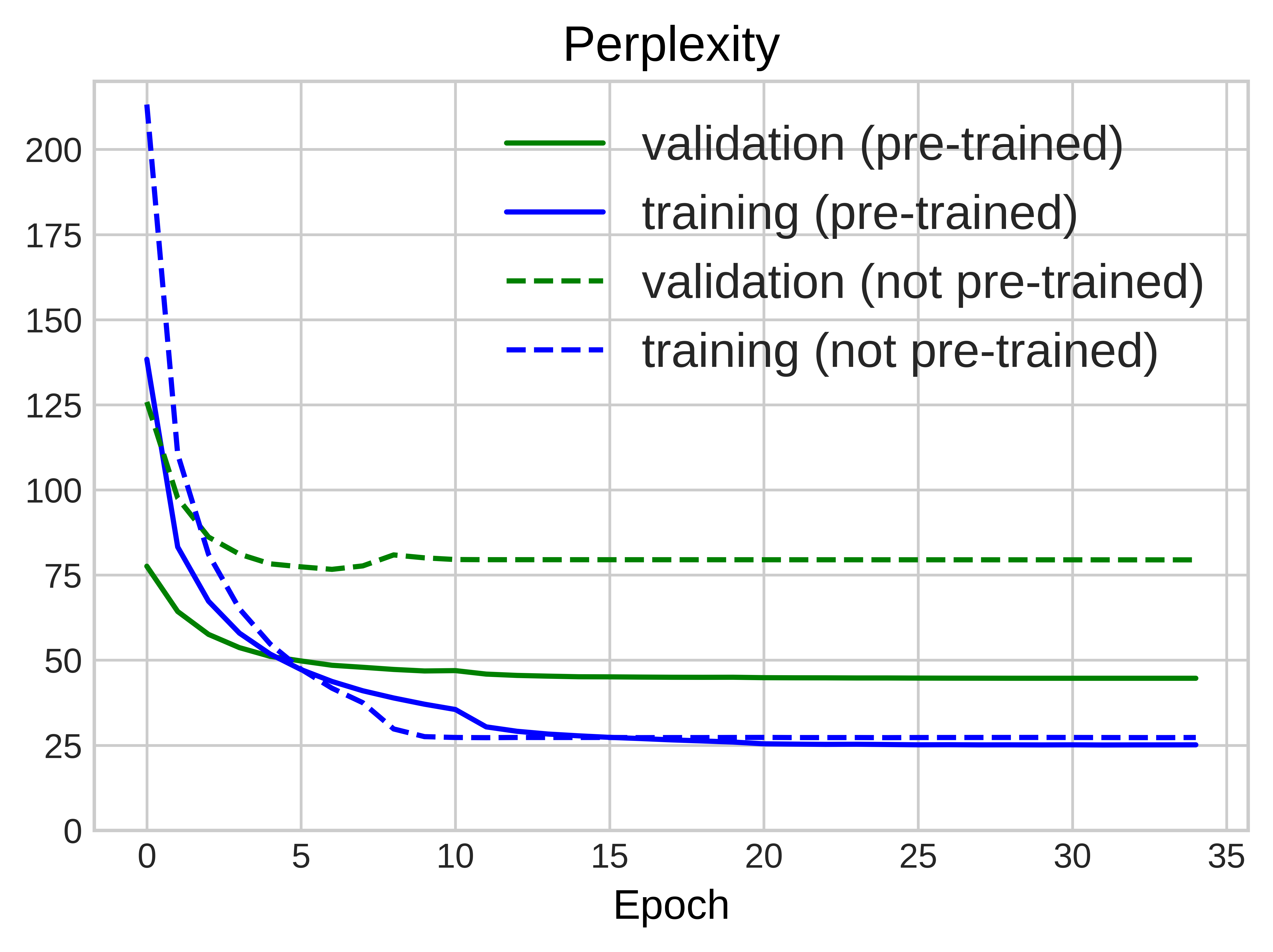}
  \caption{Perplexities of \textit{S+I Attention} model for the case of pre-training (solid lines) and for the case when the model is trained from scratch (dashed lines).}
  \label{pretraining}
\end{figure}

It is well-known that good language model can often improve metrics such as BLEU for a particular NLP task \cite{jozefowicz2016exploring}. According to this, we decided to pre-train our models. For this purpose, WikiText-103 dataset \citep{merity2016pointer} was chosen. During pre-training we set $v^{*}$ (eq. \ref{eq:5}) to zero vector to make our models purely unconditional. Embeddings for these language models were initialized by Google Word2Vec vectors\footnote{https://code.google.com/archive/p/word2vec/} and were fine-tuned. Figure \ref{pretraining} shows that this procedure helps to decrease perplexity and prevents over-fitting. Attention Skip-gram vectors were also trained on the WikiText-103.

\subsection{Results}

\begin{table*}[ht]
  \centering
    \begin{tabular}{  c ||  l  || l  }
     \hline
     Word & Context & Definition\\
     \hline
     star & she got star treatment & a person who is very important \\
     \hline
     star & bright star in the sky & \makecell[l]{a small circle of a celestial object \\ or planet that is seen in a circle} \\
     \hline
     sentence & sentence in prison & an act of restraining someone or something \\
     \hline
     sentence & write up the sentence & a piece of text written to be printed \\
     \hline
     head & the head of a man & the upper part of a human body \\
     \hline
     head & he will be the head of the office & the chief part of an organization, institution, etc \\
     \hline
     reprint & \makecell[l]{they never reprinted the \\ famous treatise} & \makecell[l]{a written or printed version of \\ a book or other publication} \\
     \hline
     rape & \makecell[l]{the woman was raped on \\ her way home at night} & the act of killing \\
     \hline
     invisible & \makecell[l]{he pushed the string through \\ an inconspicuous hole} & not able to be seen \\
     \hline
     shake & my faith has been shaken & cause to be unable to think clearly \\
     \hline
     nickname & \makecell[l]{the nickname for the u.s. \\ constitution is `old ironsides '} & a name for a person or thing that is not genuine \\
     \hline
    \end{tabular}
  \caption{Examples of definitions generated by \textit{S + I-Attention} model for the words and contexts from the test set.}
  \label{table:defsgenerated}
\end{table*}

\begin{table}[ht]
    \centering
    \begin{tabular}{  l ||  c  | c  }
     \hline
     Model & PPL & BLEU\\
     \hline
     \textit{S+G+CH+HE} (1)   & 45.62 & 11.62 $\pm$ 0.05 \\
     \textit{S+G+CH+HE} (2)  & 46.12 & -\\
     \textit{S+G+CH+HE} (3)  & 46.80 & -\\
     \textit{S\ +\ I-Adaptive} (2) &  46.08 & 11.53 $\pm$ 0.03 \\
     \textit{S\ +\ I-Adaptive} (3) &  46.93 & -\\
     \textit{S\ +\ I-Attention} (2) &  \textbf{43.54} & \textbf{12.08 $\pm$ 0.02} \\
     \textit{S\ +\ I-Attention} (3) &  44.9 & -\\
     \hline
    \end{tabular}
    \captionof{table}{Performance comparison between best model proposed by Noraset et al. \shortcite{noraset2016definition} and our models on the test set. Number in brackets means number of LSTM layers. BLEU is averaged across 3 trials.}
    \label{table:perplexity}
\end{table}

Both our models are LSTM networks \cite{Hochreiter:1997:LSM:1246443.1246450} with an embedding layer. The attention-based model has own embedding layer, mapping context words to vector representations. Firstly, we pre-train our models using the procedure, described above. Then, we train them on the collected dataset maximizing log-likelihood objective using Adam \cite{DBLP:journals/corr/KingmaB14}. Also, we anneal learning rate by a factor of 10 if validation loss doesn't decrease per epochs. We use original Adaptive Skip-gram vectors as inputs to \textit{S+I-Adaptive}, which were obtained from the official repository\footnote{https://github.com/sbos/AdaGram.jl}.
We compare different models using perplexity and BLEU score on the test set. BLEU score is computed only for models with the lowest perplexity and only on the test words that have multiple meanings. The results are presented in Table \ref{table:perplexity}. We see that both models that utilize knowledge about meaning of the word have better performance than the competing one. 
We generated definitions using \textit{S + I-Attention} model with simple temperature sampling algorithm ($\tau=0.1$). Table \ref{table:defsgenerated} shows the examples. The source code and dataset will be freely available \footnote{https://github.com/agadetsky/pytorch-definitions}.
\section{Conclusion}
In the paper, we proposed two definition models which can work with polysemantic words. We evaluate them using perplexity and measure the definition generation accuracy with BLEU score. Obtained results show that incorporating information about word senses leads to improved metrics. Moreover, generated definitions show that even implicit word context can help to differ word meanings. In future work, we plan to explore individual components of word embedding and the mask produced by our attention-based model to get a deeper understanding of vectors representations of words. 
\section*{Acknowledgments}
This work was partly supported by Samsung Research, Samsung Electronics, Sberbank AI Lab and the Russian Science Foundation grant 17-71-20072. 

\bibliography{acl2018}
\bibliographystyle{acl_natbib}


\end{document}